\documentclass{article} 
\usepackage{iclr2027_conference,times}

\usepackage{amsmath,amsfonts,bm}

\def\eqref#1{equation~\ref{#1}}

\def\1{\bm{1}}

\DeclareMathAlphabet{\mathsfit}{\encodingdefault}{\sfdefault}{m}{sl}
\SetMathAlphabet{\mathsfit}{bold}{\encodingdefault}{\sfdefault}{bx}{n}

\usepackage{hyperref}
\usepackage{url}
\usepackage{adjustbox}
\usepackage{bm}
\usepackage{multirow}
\usepackage{array}
\usepackage{makecell}   
\usepackage{graphicx}
\usepackage{amsfonts}   
\usepackage{amsmath}    
\usepackage{booktabs}   
\usepackage[table,xcdraw]{xcolor}   
\usepackage{color}    
\usepackage{xcolor}
\usepackage{pifont}
\definecolor{lightblue}{RGB}{230, 245, 255}
\definecolor{lightgreen}{RGB}{235, 250, 235}
\usepackage{algorithm}
\usepackage{algorithmic}
\usepackage{setspace}
\usepackage{multirow}
\usepackage{tikz}
\usepackage{pgfplots}
\usepackage{amssymb}
\usepackage{adjustbox}
\usepackage{amsmath}
\usepackage{float}
\usepackage{caption}
\usepackage{wrapfig}

\title{SPOON: Towards Coherent Compositional \\ 3D Scene Generation from Uncalibrated \\ Multi-view Images}

\author{%
{\bfseries
Guibiao Liao$^{1,2,*}$,\hspace{0.5em}
Mochu Xiang$^{1,2,5,*}$,\hspace{0.5em}
Heng Li$^{3,2}$,\hspace{0.5em}
Ken Deng$^{3,2}$,\hspace{0.5em}
Zijie Wang$^{4,2}$,}\\
{\bfseries
Guanbin Li$^{4,2}$,\hspace{0.5em}
Ping Tan$^{3,2}$,\hspace{0.5em}
Shenghua Gao$^{1,2,5}$,\hspace{0.5em}
Yizhou Yu$^{1,2}$}
}

\iclrfinalcopy 
\begin{document}

\maketitle
\lhead{Preprint}

\begingroup
\makeatletter
\renewcommand{\@makefntext}[1]{%
  \noindent #1%
}
\makeatother

\footnotetext{%
$^{1}$The University of Hong Kong
\quad
$^{2}$Shenzhen Loop Area Institute
\quad
$^{3}$The Hong Kong University of Science and Technology
\quad
$^{4}$Sun Yat-sen University.
\quad
$^{5}$TranscEngram.
\quad
$^{*}$Equal contribution.
}
\endgroup

\begin{figure}[h]
    \centering
    \includegraphics[width=.96\linewidth]{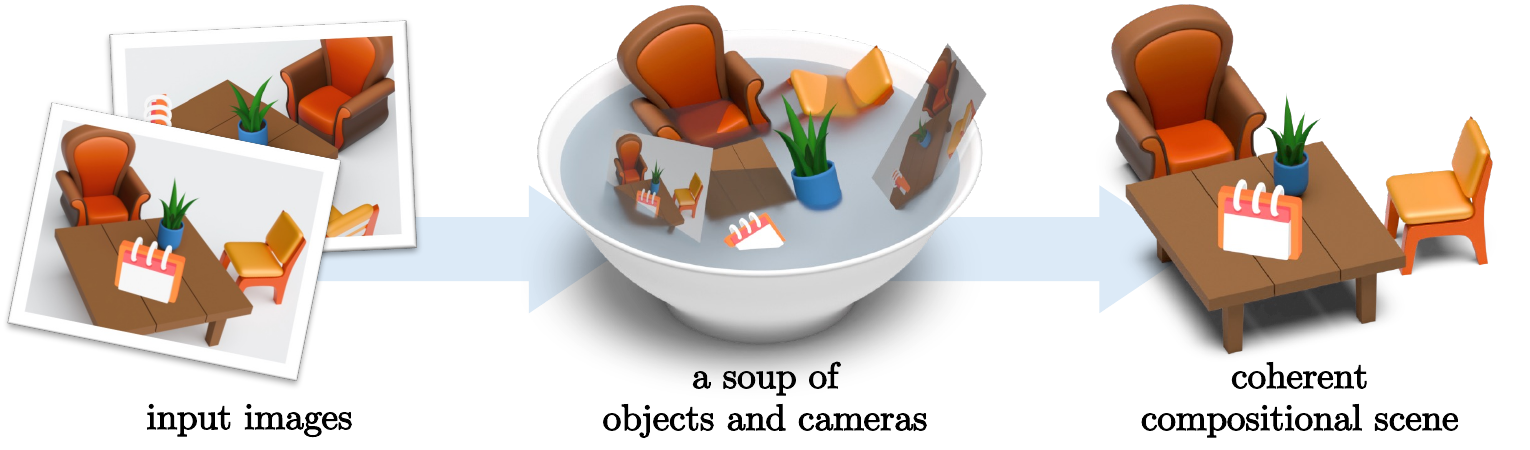}
    \vspace{-3mm}
    \caption{Given multi-view images, current 3D generation pipelines often yield a soup of objects. SPOON can recover a coherent compositional 3D scene with accurate object geometry and layout.}
    \label{fig:teaser_image}
\end{figure}

\begin{abstract}

Compositional 3D scene generation aims to recover complete 3D object shapes and their spatial arrangement from visual observations. Recent image-conditioned 3D generators provide strong priors for producing high-quality object geometry, making the generation of complex scenes increasingly practical. A central challenge is therefore to spatially organize these generated assets into a globally coherent scene while remaining consistent with multi-view observations. 
Existing approaches either entangle scene layout with object generation or separately estimate spatial placement from view-specific observations, where pose hypotheses may remain ambiguous and inconsistent across views, often resulting in an incoherent \emph{object-camera soup}. 
We introduce \textsc{SPOON}, a framework that reformulates multi-view compositional 3D generation as scene-level, geometry-grounded pose reasoning. 
Rather than treating view-specific object pose hypotheses independently, \textsc{SPOON} coordinates them using reconstruction-derived multi-view geometry through a \emph{Guide-Route-Reconcile} paradigm.  
This progressively organizes object poses and camera configurations into a coherent scene-level spatial arrangement. 
Extensive experiments on ARSG-110K and MIDI-3D-Front demonstrate consistent improvements in object placement and scene composition across varying numbers of input views. On ARSG-110K, \textsc{SPOON} reduces scene-level and object-level Chamfer distances by $12.7\%$ and $17.7\%$, respectively, compared with a strong baseline.

\end{abstract}
    
\section{Introduction}
\label{sec:intro}

Multi-view compositional 3D scene generation can construct coherent scenes by recovering complete object assets together with their spatial arrangement from visual observations, supporting applications in embodied AI, simulation, and 3D content creation.
Recent advances in image-conditioned 3D generation \cite{trellis, trellis2, cupid, sam3d} provide strong priors for recovering high-quality object geometry, making compositional 3D scene generation increasingly practical. 
A challenge is therefore how to incorporate multi-view observations and recover the spatial layout that organizes these generated assets into a coherent scene.

One strategy is to incorporate layout directly into the generation process.
For example, 3D-Fixer \cite{3dfixer} conditions object completion on partial geometry in the scene coordinate frame, using the observed geometry as a spatial anchor to generate objects in place.
While avoiding explicit post-generation pose recovery, this formulation entangles shape generation with scene-space placement, requiring object generative priors learned in canonical space to additionally accommodate incomplete observations under arbitrary spatial transformations. 

An alternative strategy preserves object generation in its canonical space and models spatial placement separately.
SAM 3D \cite{sam3d}, for example, generates canonical object geometry while recovering its transformation with respect to the observing camera.
Such a decomposition preserves the advantages of object-centric generative priors, while making accurate pose recovery critical to coherent scene composition.
Pose estimates inferred from individual object-view observations are inherently scale-ambiguous and can become inaccurate under occlusion, image-boundary truncation, and limited visibility.
Multiple views provide complementary evidence, yet do not automatically resolve this ambiguity: different observations may support pose estimation with substantially different reliability, and independently plausible hypotheses can become mutually inconsistent when expressed in a shared scene frame.
Thus, these methods often provide an incoherent \emph{object--camera soup}.

Scene-level 3D reconstruction provides complementary geometric cues for resolving this ambiguity.
It jointly recovers camera geometry and scene structure in a shared coordinate system, establishing globally consistent relations across observations.
This shared geometry provides a common reference for coordinating object hypotheses across cameras and reasoning about their support within the overall scene.
This raises a natural question:
\emph{Can scene-level geometry organize locally plausible object–camera hypotheses into a globally coherent 
3D composition?
}

We answer this question with \textsc{SPOON}, standing for \textbf{S}cene-level \textbf{P}ose and \textbf{O}bject layout \textbf{O}ptimizatio\textbf{N}.
We cast multi-view object placement as a \emph{scene-level pose reasoning problem}, where view-conditioned hypotheses are coordinated through shared scene geometry rather than considered in isolation.
In this way, \textsc{SPOON} \emph{stirs the object-camera soup into a coherent scene}, combining strong object-centric generative priors with scene-level geometric reasoning for coherent spatial placement. 
\textsc{SPOON} follows a \emph{Guide--Route--Reconcile} paradigm.
First, we \textbf{Guide} pose generation with reconstruction-derived scene geometry, grounding view-conditioned predictions in globally consistent spatial cues.
Second, we \textbf{Route} the resulting pose hypotheses by assessing the geometric reliability of each view-specific observation against the reconstructed object geometry.
Finally, we \textbf{Reconcile} the object layouts and camera poses through joint differentiable optimization, reducing residual object--camera inconsistencies while preserving the generated object geometry. 
Together, these stages progressively coordinate local pose hypotheses into a coherent scene-level spatial configuration without altering the underlying object generative prior. 
Our main contributions are:

\begin{itemize}

\item We identify the \emph{object--camera soup} problem in compositional 3D scene generation, where locally plausible object-camera hypotheses may not constitute a coherent scene-level configuration.

\item We propose \textsc{SPOON}, a scene-level pose reasoning framework that follows a \emph{Guide--Route--Reconcile} paradigm to guide pose generation using shared geometry, route view-specific hypotheses by geometric reliability, and jointly reconcile object layouts and camera poses.

\item Extensive experiments on ARSG-110K and MIDI-3D-Front show consistent improvements in scene composition and object placement across varying numbers of input views. 

\end{itemize}

\section{Related work}
\label{sec:related_work}

\subsection{3D Shape Generation}

Recent advances in 3D generation have substantially improved the quality and scalability of object-level asset creation from textual or visual inputs. 
Early methods were largely developed on relatively small-scale 3D datasets \cite{pixel2mesh,meshrcnn,shapenet}, whereas the emergence of large-scale collections such as Objaverse \cite{objaverse} and Objaverse-XL \cite{objaversexl} has enabled increasingly powerful generative priors. 
Modern native 3D generators explore a wide range of representations, including VecSets \cite{3dshape2vecset}, tri-planes \cite{get3d,direct3d}, geometry images \cite{omage,geometryimagediffusion}, 3D primitives \cite{3dtopiaxl}, structured latents \cite{trellis,direct3ds2,ultra3d}, and autoregressive mesh representations \cite{meshllm,meshxl,edgerunner,meshgpt,meshanything,meshanythingv2}. 
Together with advances in scalable architectures and optimized 3D operators \cite{direct3ds2,ultra3d,trellis2}, these developments have significantly improved the geometric fidelity and structural complexity of generated 3D assets. 
The increasing availability of high-quality object generation makes it increasingly feasible to extend object-level generative models toward compositional 3D scene generation. 
Our work builds upon this progress and focuses on reliable object placement and spatial organization in complex multi-view scenes.

\subsection{Compositional 3D Scene Generation}
Building upon strong object-level generative priors, recent studies have extended 3D generation toward compositional scenes with multiple objects and observations.
Two design questions are particularly relevant in this setting: how to integrate complementary evidence across views, and how to recover the spatial layout of generated objects within a shared scene frame.

For multi-view object generation, recent approaches increasingly combine reconstruction and generative priors to improve geometric completeness and input consistency.
ReconViaGen \cite{reconviagen}, UniRecGen \cite{unirecgen}, and Mix3R \cite{mix3r} transfer knowledge from 3D reconstruction foundation models \cite{vggt, pi3} to condition 3D generators on multi-view observations.  
These approaches demonstrate the benefit of integrating reconstruction priors or multi-view consistency into generative models, but primarily focus on aggregating information across observations for object shape enhancement.

Accurately modeling object layouts further requires 3D generation models to be explicitly pose-aware. 
Recent approaches decouple shape from camera or object pose using representations such as UV volumes \cite{cupid} or camera tokens \cite{sam3d}. 
For example, ShapeR \cite{shaper} first predicts object-to-world transformations from detected bounding boxes and subsequently generates their 3D shapes under the estimated placements. 
Another line of work directly applies layout transformations to object representations and learns to model the resulting SE(3)-transformed shapes within a scene volume. While this formulation enables pixel-aligned geometry \cite{pixal3d}, entangling arbitrary object transformations with shape representation broadens the distribution that the generative model must capture and can degrade the geometry of generated meshes \cite{iscene, 3dfixer}. 
This motivates compositional formulations that preserve object geometry in canonical space while modeling spatial transformations separately. 
SceneGen \cite{scenegen} predicts object positions before assembling independently generated assets, while CAST \cite{cast} further refines the resulting scene composition using photometric and physics-aware objectives. 
MV-SAM3D \cite{mvsam3d} extends to multi-view observations through Multi-Diffusion \cite{multidiffusion} and confidence-aware latent velocity fusion. 
These formulations retain strong object-level generative priors while exposing spatial layout as an explicit variable for scene composition.

\section{Preliminaries}
\label{sec:preliminaries}

\subsection{Joint Shape and Pose Modeling} 
Given an image prompt, \textsc{SAM3D} \cite{sam3d} can recover the object shape and the camera pose. It adopts a two-stage flow-matching architecture: the first stage jointly generates canonical object geometry (as low-resolution occupancy) and the relative camera pose; the second stage generates the object's texture and fine geometry. 
Given an object observation $\mathcal{I}$, the first-stage shape and pose branches predict the corresponding velocity fields: 
\begin{equation}
\mathbf{v}^{S}_{\theta}, \mathbf{v}^{P}_{\theta}=f_{\theta}(\mathbf{z}^{S}_{t},\mathbf{z}^{P}_{t},\mathcal{I},t),
\label{eq:joint_shape_pose_flow}
\end{equation}
where $\mathbf{z}^{S}_{t}$ denotes the shape latent and $\mathbf{z}^{P}_{t}=(\mathbf{z}^{R}_{t},\mathbf{z}^{T}_{t},\mathbf{z}^{s}_{t})$ is the camera-relative pose latent, parameterizing rotation, translation, and scale at flow time $t$. 
Under rectified flow, both states evolve according to their predicted velocities:
\begin{equation}
\mathbf{z}^{S}_{t+\Delta t}=\mathbf{z}^{S}_{t}+\Delta t\,\mathbf{v}^{S}_{\theta}, \qquad
\mathbf{z}^{P}_{t+\Delta t}=\mathbf{z}^{P}_{t}+\Delta t\,\mathbf{v}^{P}_{\theta} .
\label{eq:joint_flow_update}
\end{equation}
At the endpoint, the shape latent is decoded into a canonical object representation $\mathcal{G}_o$, while the pose state yields a camera-frame transformation $T^{c}_{o,v}$ that places the generated object relative to view $v$. 
Instead of entangling object shape with its layout, separately modeling geometry and layout keeps the data distribution simple and easy to learn, which is beneficial in generating high-fidelity 3D shapes. However, this might sacrifice the accuracy of the layout estimation.

\subsection{Multi-View Flow Fusion}
\label{sec:multi_diffusion}
Multi-Diffusion \cite{multidiffusion} fuses prompt-conditioned velocity predictions into a global coherent direction at each generation step, which makes a flow matching model accept multiple conditions. 
MV-\textsc{SAM3D} \cite{mvsam3d} extends \textsc{SAM3D} to multi-view generation to exploit complementary cues across observations. 
Given $V$ views, the corresponding shape velocities $\{\mathbf{v}^{S}_{\theta,v}\}_{v=1}^{V}$ are aggregated into a shared update: 
\begin{equation}
\mathbf{v}^{S}_{\theta}=\mathcal{A}\!\left(\{\mathbf{v}^{S}_{\theta,v}\}_{v=1}^{V}\right), \qquad
\mathbf{z}^{S}_{t^+}=\mathbf{z}^{S}_{t}+\Delta t\,\mathbf{v}^{S}_{\theta},
\label{eq:multiview_shape_fusion}
\end{equation}
where $\mathcal{A}$ denotes the multi-view velocity fusion operator, $t^+=t+\Delta t$.
This yields a shared shape trajectory that integrates complementary evidence across observations. 
In contrast, the pose trajectories remain view-conditioned and evolve independently. 
Consequently, multi-view fusion produces a shared canonical shape while retaining view-specific camera-relative pose hypotheses.

\section{Method}
\label{sec:method}

Given $V$ unposed RGB images $\{I_v\}_{v=1}^{V}$ and the corresponding instance masks $\{M_{o,v}\}$, our goal is to reconstruct a compositional 3D scene containing $O$ observed objects.
We leverage feedforward 3D reconstruction models~\cite{vggtomega} to recover per-view point maps $X_v$ and camera intrinsics $K_v$ and world-to-camera extrinsics $E_v$ in a shared world coordinate system. 

For each object, we follow MV-\textsc{SAM3D} \cite{mvsam3d} to obtain a shared canonical representation $\mathcal{G}_o$ together with view-specific camera-frame pose hypotheses $\{T^{c}_{o,v}\}_{v=1}^{V}$. 
We seek an object-to-world transformation $T_o^{\mathrm{ow}}$ for each canonical object, yielding the compositional scene: 
\begin{equation}
\mathcal{S}=\left\{\left(\mathcal{G}_o,T_o^{\mathrm{ow}}\right)\right\}_{o=1}^{O}, 
\label{eq:scene_representation}
\end{equation}
where $T_o^{\mathrm{ow}}$ places $\mathcal{G}_o$ into the shared world frame.

\begin{figure}[t]
    \centering
    \includegraphics[width=.99\linewidth]{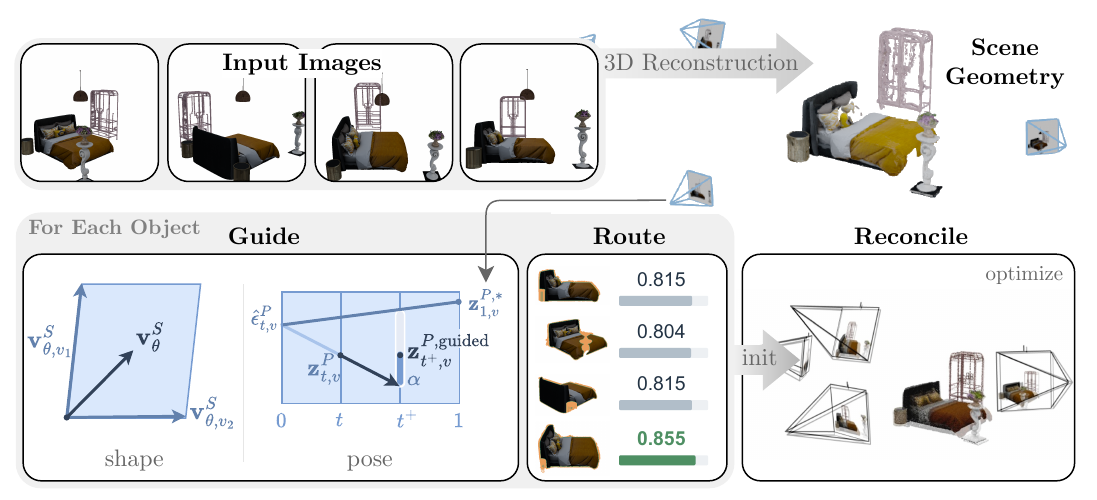}
    \vspace{-2mm}
    \caption{\textbf{Overview of SPOON.} 
    Given multiple uncalibrated images, 3D reconstruction methods first recover the scene geometry as camera poses and depth maps. Then, for each object, we prompt the mesh generation process with multi-view visual cues and \textit{guide} the pose recovery with reconstruction guidance, which is then placed using an optimal layout hypothesis through \textit{routing}. 
    Finally, the object layouts and camera poses are jointly optimized and \textit{reconciled} to provide a coherent compositional 3D scene.
    }
    \label{fig:pipeline_image}
\end{figure}

While MV-\textsc{SAM3D} produces plausible camera-relative object placements, these view-specific hypotheses may collectively lack a scene-coherent spatial configuration, giving rise to an \emph{object--camera soup}. 
To address this, \textsc{SPOON} uses multi-view geometric cues as a scene-level reference and organizes these hypotheses through a \emph{Guide--Route--Reconcile} paradigm, as described below.

\subsection{Geometry-Guided Object Pose Generation}
\label{sec:pose_guidance}
While multi-view flow fusion (Sec. \ref{sec:multi_diffusion}) aggregates complementary cues for shape generation, pose generation in \cite{mvsam3d} remains view-specific.
Such independently inferred poses may be ambiguous and unreliable from a single image, whereas feed-forward 3D reconstruction models \cite{vggtomega} recover camera poses from the full multi-view input with stronger geometric consistency. 
Thus, we propose to use these reconstruction-derived poses as an inference-time geometric prior to guide pose generation.

A straightforward solution is to align reconstruction-derived camera poses in the world frame with generated poses in the object canonical frame and use the aligned poses as guidance.
However, this requires decoding intermediate pose states from the flow trajectory, which are often unstable and can introduce substantial errors into the cross-frame alignment.
In contrast, the canonical object surface remains substantially more stable than the decoded pose states and therefore provides a more reliable anchor for cross-frame alignment. 
Therefore, we establish the cross-frame correspondence through object geometry, aligning the aggregated reconstruction with the canonical object surface to bridge the two coordinate systems.

\textbf{Surface coordinate bridge.}
Concretely, we construct object surfaces in both coordinate frames. 
Using the point maps and camera extrinsics from 3D reconstruction methods~\cite{vggtomega}, we transform the masked object points into the shared world frame and aggregate them into a world-space point cloud $\mathcal{P}_{W}$.  
To establish a correspondence with the canonical frame where each shape is generated, we recover a clean occupancy from the noisy shape latent and the fused velocity:
\begin{equation}
    \widehat{\mathbf{z}}^{S}_{1}
    =
    \mathbf{z}^{S}_{t}
    +
    (1-t)\mathbf{v}^{S}_{\theta},
    \label{eq:shape-endpoint}
\end{equation}
where $\mathbf{v}^{S}_{\theta}$ denotes the fused shape velocity. 
We decode $\widehat{\mathbf{z}}^{S}_{1}$ and sample its surface to obtain the canonical point cloud $\mathcal{P}_{C}$. 
Since $\mathcal{P}_{W}$ and $\mathcal{P}_{C}$ lie in different coordinate systems, solving the transformation between them can robustly transfer the reconstructed camera pose into the canonical space in shape generation. 
After centering and normalizing both point cloud sets, we search for an optimal rotation that best aligns them, yielding the bridge-aligned pose $\boldsymbol{\pi}^{*}_{v}$. 
We then encode this aligned pose using the \textsc{SAM3D} pose encoder: 
\begin{equation}
\mathbf{z}^{P,*}_{1,v} = \operatorname{Enc}_{P}\left(\boldsymbol{\pi}^{*}_{v}\right),
\label{eq:geometry_pose_endpoint}
\end{equation}
which serves as the geometry-derived clean endpoint for the pose rectified flow.

\textbf{Flow-consistent endpoint injection.}
Directly replacing the intermediate pose state $\mathbf{z}^{P}_{t,v}$ with the geometry-derived endpoint $\mathbf{z}^{P,*}_{1,v}$ would ignore the current flow time and deviate from the learned trajectory. 
We instead construct a reference state at the same flow time. 
From the current state and predicted velocity, we recover the corresponding noise endpoint as $\widehat{\boldsymbol{\epsilon}}^{P}_{t,v}=\mathbf{z}^{P}_{t,v}-t\mathbf{v}^{P}_{\theta,v}$. 
We then keep this noise realization fixed and replace only the clean endpoint with $\mathbf{z}^{P,*}_{1,v}$ to define a time-matched reference state, which is then used to guide the Euler update:
\begin{align}
\mathbf{z}^{P,\mathrm{ref}}_{t^+,v}
& =
(1-t^+)\widehat{\boldsymbol{\epsilon}}^{P}_{t,v} +
t^+\mathbf{z}^{P,*}_{1,v},
\label{eq:reference-flow-state}\\
\mathbf{z}^{P,\mathrm{guided}}_{t^+,v}
&=
(1-\alpha_P)\mathbf{z}^{P,\mathrm{base}}_{t^+,v} +
\alpha_P\mathbf{z}^{P,\mathrm{ref}}_{t^+,v}.
\label{eq:guided-pose-flow}
\end{align}
where $\mathbf{z}^{P,\mathrm{base}}_{t^+,v}$ denotes the standard Euler proposal.
This update retains the noise endpoint inferred from the current linear flow approximation while steering the clean endpoint toward the geometry-derived pose.

\subsection{Geometry-Aware Pose Routing}
\label{sec:pose_routing} 
Geometry guidance improves object pose generation, yet the resulting hypotheses can still vary in reliability across views.
Differences in visibility and boundary truncation may affect the geometric evidence available for pose estimation.
Since pose accuracy cannot be directly assessed at inference time, we use geometric observability as a proxy for reliability.
Specifically, we derive a per-view reliability score from its support under the aggregated multi-view geometry and use it to select the pose hypothesis for world-frame placement.
For clarity, we omit the object index $o$ below.

\textbf{Multi-view geometric support.}
We first lift the masked point-map observations from all views into the shared world frame using the estimated camera extrinsics.
Their union is voxel-downsampled into an aggregated object point cloud $\mathcal{P}_{W}$, which combines complementary observations and serves as a geometric proxy for the object extent.
For each view $v$, we reproject $\mathcal{P}_{W}$ onto the image plane: 
$
\widehat M_v = \operatorname{Rasterize}
\left( \left\{
\Pi\!\left( K_v\,\mathcal{T}(E_v,\mathbf P) \right)
\;\middle|\;
\mathbf P\in \mathcal{P}_{W} \right\} \right),
$
where $\Pi(\cdot)$ denotes perspective projection.  
We evaluate $\widehat M_v$ before restricting it to the valid image domain $\Omega_v$, so that it also captures geometry projected beyond the image boundary. 
Together with the observed mask $M_v$, this provides complementary cues for boundary truncation and visible coverage.

\textbf{Geometry-aware reliability.}
We characterize geometric observability using frame completeness $F_v$ and visible coverage $C_v$: 
\begin{equation}
\begin{array}{ccc}
F_v = \displaystyle\frac{|\widehat M_v \cap \Omega_v|} {|\widehat M_v|},
&
C_v = \displaystyle\frac{|M_v \cap \widehat M_v|} {|\widehat M_v \cap \Omega_v|},
 & 
S_v = \left[F_v \geq \tau_f\right] F_v^{\alpha} C_v, 
\end{array}
\label{eq:reliability_score}
\end{equation}
where $F_v$ measures how much of the projected object extent remains inside the image, while $C_v$ measures how much of the in-frame geometric support is covered by the observation.
Here, $\tau_f$ is the completeness threshold used to filter severely truncated observations. 
Their combination $S_v$ serves as a geometry-aware proxy for pose reliability. 

\textbf{Reliability-aware pose routing.}
We select the most reliable view and transfer its pose hypothesis to the shared world frame:
\begin{equation}
    v^{\star} = \operatorname*{arg\,max}_{v} S_v,
    \qquad
    T_o^{\mathrm{ow}} = E_{v^{\star}}^{-1} T_{o,v^{\star}}^{c},
    \label{eq:pose_routing}
\end{equation}
where $T_{o,v^{\star}}^{c}$ is the object transformation predicted in the selected camera frame.
This converts heterogeneous multi-view pose hypotheses into a single placement supported by the most geometrically reliable observation.

\subsection{Multi-View Pose Reconciliation}
\label{sec:pose_ba}

The strategy introduced in Sec.~\ref{sec:pose_routing} improves the object layout by selecting the most plausible layout among the candidates estimated from different input views. However, the quality of the resulting layout remains bounded by the accuracy of the initial per-view layout estimations. To further improve the consistency of the reconstructed scene, we leverage the multi-view observations and jointly refine the object layouts through differentiable Gaussian splatting.

Specifically, for the $o$-th object, we introduce an additional $\mathrm{Sim}(3)$ transformation parameterized by $(s',R',t')$ to refine its initial layout $(s,R,t)=T_o^{\mathrm{ow}}$. The resulting transformation is given by: 
\begin{equation}
s \leftarrow s's,\qquad
R \leftarrow R'R,\qquad
t \leftarrow s'R't + t'.
\end{equation}
We initialize the additional transformation as the identity, i.e., $s'=1$, $R'=I_{3\times3}$, and $t'=0$. The transformed object Gaussians are then composited and rendered from the input viewpoints, whose camera poses are obtained from a 3D reconstruction foundation model \cite{vggtomega}. 
In this way, the multi-view observations provide image-space supervision for refining the object layout without requiring explicit cross-view point or feature correspondences.

Although this optimization can effectively correct moderate layout errors, its performance remains coupled with the estimated camera geometry. 
Errors in the estimated camera poses perturb multi-view consistency, 
limiting the effectiveness of object-level refinement. 
To address this issue, we refine the camera poses jointly with the object layouts. We employ a Gaussian rasterization kernel with explicit camera Jacobians~\cite{gsslam}, which enables efficient gradient-based optimization of the camera parameters through the rendering process. Conceptually, this optimization is analogous to Bundle Adjustment (BA) in Structure-from-Motion, where camera poses and scene geometry are jointly optimized to achieve multi-view consistency. Rather than relying on explicit point or feature correspondences, our formulation directly minimizes the discrepancy between the rendered and observed images. Moreover, the scene geometry is represented by the generated object-level Gaussian representations, while the object motion is parameterized by a compact $\mathrm{Sim}(3)$ transformation.

Formally, given the set of input images, we jointly optimize the camera poses $\{E_v\}$ and the per-object similarity transformations $\{(s_o',R_o',t_o')\}$ by minimizing the multi-view rendering objective:
\begin{equation}
\min_{\{E_v\},\{s_o',R_o',t_o'\}}
\sum_{v}\mathcal{L}_{\mathrm{render}}
\left(
\mathcal{R}\left(
{\mathcal{G}_o(s_o',R_o',t_o')}_o, 
E_v
\right),
I_v
\right),
\end{equation}
where $\mathcal{G}_o$ denotes the Gaussian representation of the $o$-th object, $\mathcal{R}$ denotes differentiable Gaussian rasterization, and $I_v$ is the $v$-th input image. The rendering loss can incorporate both photometric and depth supervision, where depth is from 3D reconstruction model as a regularization term. 
By jointly optimizing the camera and object parameters, the method allows their relative configurations to adapt under shared multi-view supervision, leading to a more globally consistent reconstruction.

\begin{table*}[t]
\centering
\caption{
\textbf{Main results on the ARSG-110K and MIDI-3D-Front test sets under varying numbers of input views.}
We report scene- and object-level metrics together with 3D bounding-box IoU for spatial layout evaluation. 
Best results within each input-view setting are highlighted in bold. 
}
\vspace{-3mm}
\label{tab:main_results}
\small
\renewcommand{\arraystretch}{1.05}
\setlength{\tabcolsep}{3.pt}

\begin{tabular*}{\textwidth}{
@{\extracolsep{\fill}}
l c
ccccc
ccccc
@{}
}
\toprule

\multirow{2}{*}{Method}
& \multirow{2}{*}{Views}
& \multicolumn{5}{c}{ARSG-110K}
& \multicolumn{5}{c}{MIDI-3D-Front} \\

\cmidrule(lr){3-7}
\cmidrule(lr){8-12}

&
& $\mathrm{CD}_{S}\downarrow$
& $\mathrm{FS}_{S}\uparrow$
& $\mathrm{CD}_{O}\downarrow$
& $\mathrm{FS}_{O}\uparrow$
& $\mathrm{IoU}\uparrow$
& $\mathrm{CD}_{S}\downarrow$
& $\mathrm{FS}_{S}\uparrow$
& $\mathrm{CD}_{O}\downarrow$
& $\mathrm{FS}_{O}\uparrow$
& $\mathrm{IoU}\uparrow$ \\

\midrule
Gen3DSR
& 1
& 0.265 & 46.72 & 0.546 & 31.95 & 0.304
& 0.123 & 40.07 & 0.157 & 38.11 & 0.363 \\

MIDI
& 1
& 0.801 & 15.35 & \textbf{0.179} & 35.99 & 0.033
& 0.080 & 50.19 & 0.103 & 53.58 & \textbf{0.518} \\

I-Scene
& 1
& 0.277 & 34.69 & 0.563 & 21.25 & 0.228
& 0.311 & 10.76 & 0.187 & 69.90 & 0.025 \\

3D-Fixer
& 1
& 0.159 & \textbf{68.82} & 0.197 & \textbf{57.85} & \textbf{0.519}
& \textbf{0.069} & \textbf{78.67} & \textbf{0.032} & \textbf{94.39} & 0.492 \\

SAM3D
& 1
& \textbf{0.146} & 65.32 & 0.257 & 48.48 & 0.445
& 0.084 & 77.06 & 0.053 & 92.62 & 0.430 \\

\midrule
ShapeR
& 4
& 0.073 & 78.10 & 0.148 & 61.83 & 0.502
& 0.082 & 87.28 & 0.024 & 96.31 & 0.506 \\

MV-SAM3D
& 4
& 0.063 & 82.37 & 0.113 & 69.48 & 0.585
& 0.083 & 87.84 & 0.020 & 96.76 & 0.597 \\

\textbf{Ours}
& 4
& \textbf{0.055} & \textbf{87.08} & \textbf{0.093} & \textbf{78.02} & \textbf{0.662}
& \textbf{0.069} & \textbf{91.73} & \textbf{0.016} & \textbf{97.66} & \textbf{0.684} \\

\midrule
MV-SAM3D
& 6
& 0.056 & 84.56 & 0.108 & 72.98 & 0.606
& 0.086 & 85.62 & 0.030 & 95.40 & 0.568 \\

\textbf{Ours}
& 6
& \textbf{0.047} & \textbf{89.68} & \textbf{0.087} & \textbf{81.42} & \textbf{0.695}
& \textbf{0.070} & \textbf{91.13} & \textbf{0.023} & \textbf{97.09} & \textbf{0.684} \\

\midrule
MV-SAM3D
& 8
& 0.051 & 86.90 & 0.101 & 74.33 & 0.615
& 0.083 & 86.23 & 0.029 & 95.54 & 0.572 \\

\textbf{Ours}
& 8
& \textbf{0.038} & \textbf{92.84} & \textbf{0.078} & \textbf{84.54} & \textbf{0.723}
& \textbf{0.069} & \textbf{91.21} & \textbf{0.026} & \textbf{96.62} & \textbf{0.684} \\

\bottomrule
\end{tabular*}
\vspace{-4mm}
\end{table*}

\section{Experiments}
\label{sec:experiments}

\subsection{Experimental Setup}
\textbf{Implementation Details.} 
We use VGGT-Omega \cite{vggtomega} to estimate camera poses and scene point cloud from the input views, and adopt SAM3D with its default settings as the underlying object generator. 
All experiments are conducted on a single NVIDIA H100 GPU. 
We evaluate our method on two benchmarks for compositional 3D scene generation: the ARSG-110K test set \cite{3dfixer} and the MIDI-3D-Front \cite{3dfront}. 
Both benchmarks provide object-level geometry together with ground-truth scene configurations, enabling evaluation of object placement and overall scene composition.

\begin{figure*}[t]
    \centering
    \includegraphics[width=\textwidth]{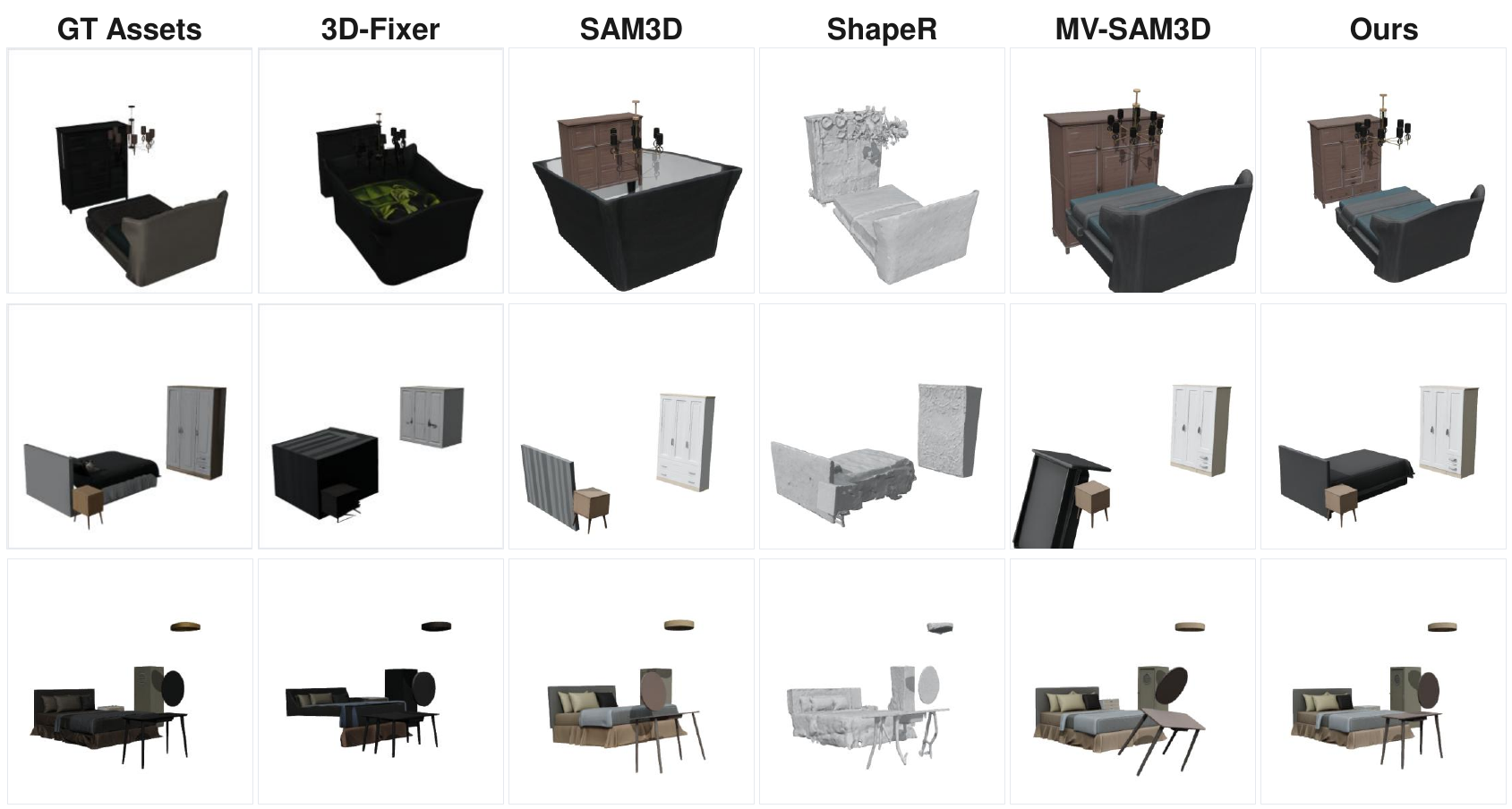}
    \vspace{-6mm}
    \caption{
    \textbf{Qualitative comparison on the MIDI-3D-Front.}
    Compared with existing single-view and multi-view methods, our method produces more accurate object placements and scene layouts.
    }
    \label{fig:qualitative_midi}
    \vspace{-2mm}
\end{figure*}

\begin{figure*}[t]
    \centering
    \includegraphics[width=\textwidth]{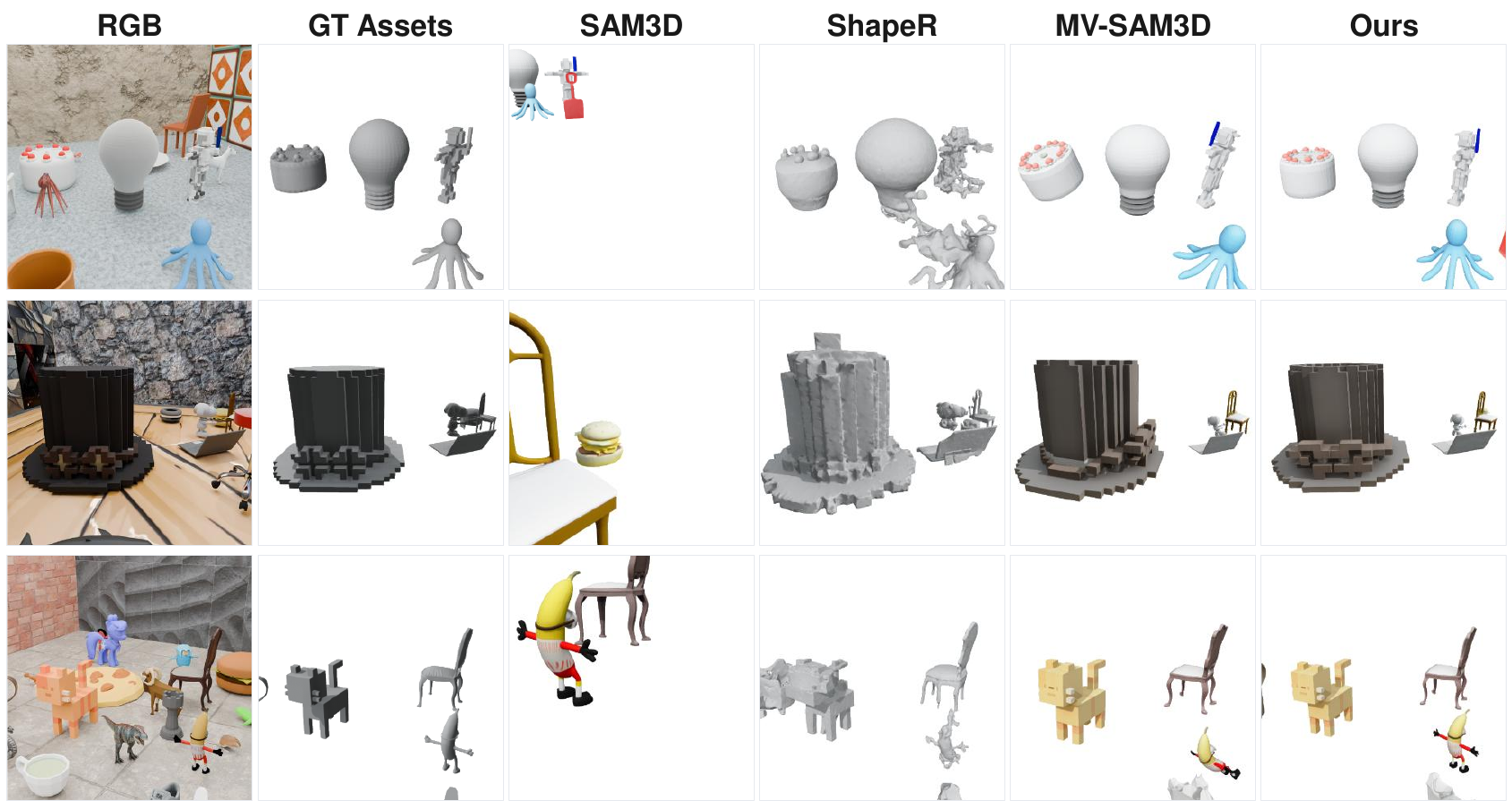}
    \vspace{-6mm}
    \caption{
    \textbf{Qualitative comparison on the ARSG-110K test set.}
    }
    \label{fig:qualitative_arsg}
    \vspace{-2mm}
\end{figure*}

\textbf{Baselines.}
We compare with representative single-view and multi-view scene generation methods. Single-view baselines include Gen3DSR~\cite{gen3dsr}, MIDI~\cite{midi}, I-Scene \cite{iscene}, 3D-Fixer~\cite{3dfixer}, and SAM3D~\cite{sam3d}. 
We further compare with ShapeR~\cite{shaper} and MV-SAM3D~\cite{mvsam3d}, which leverage multiple observations for 3D generation and provide the most relevant comparisons to our multi-view setting.

\noindent
\textbf{Metrics.}
Following 3D-Fixer~\cite{3dfixer}, we evaluate scene composition at both the scene and object levels. 
We report Chamfer Distance and F-Score for the complete scene ($\mathrm{CD}_{S}$, $\mathrm{FS}_{S}$) and for individual objects ($\mathrm{CD}_{O}$, $\mathrm{FS}_{O}$), with the predicted object poses retained during evaluation. 
These metrics therefore reflect both geometric fidelity and errors in object placement. The F-Score is computed with a distance threshold of $0.1$. 
We further report the volumetric IoU between predicted and ground-truth 3D object bounding boxes as a more direct measure of spatial layout accuracy. 
Lower CD and higher F-Score and IoU indicate better performance.

\subsection{Comparison with State-of-the-Art Methods}

\noindent
\textbf{ARSG-110K.}
In Table \ref{tab:main_results}, under the 4-view setting, multi-view methods substantially outperform single-view approaches, confirming the benefit of complementary observations. 
Compared with MV-SAM3D, our method reduces $\mathrm{CD}_{S}$ and $\mathrm{CD}_{O}$ by $12.7\%$ and $17.7\%$, while improving $\mathrm{FS}_{S}$ and $\mathrm{FS}_{O}$ by $5.7\%$ and $12.3\%$, respectively. More notably, object-layout IoU improves by $13.2\%$, showing that geometry-grounded pose reasoning more effectively converts complementary multi-view evidence into accurate object placement and coherent scene composition. 

\noindent
\textbf{MIDI-3D-Front.}
A similar trend is observed on MIDI-3D-Front. 
Under the 4-view setting, our method reduces $\mathrm{CD}_{S}$ and $\mathrm{CD}_{O}$ by $16.9\%$ and $20.0\%$ over MV-SAM3D, respectively, while improving object-layout IoU substantially by $14.6\%$. 
This suggests that the main advantage of our method lies in more reliable spatial placement and scene composition rather than merely improving object-level agreement.

\noindent
\textbf{Scaling with input views.}
Our method remains robust as the number of input views increases. 
On ARSG-110K, the relative IoU gain over MV-SAM3D increases from $13.2\%$ with 4 views to $14.7\%$ with 6 views and $17.6\%$ with 8 views, while the reduction in $\mathrm{CD}_{S}$ grows from $12.7\%$ to $16.1\%$ and $25.5\%$, respectively. 
On MIDI-3D-Front, SPOON maintains stable scene-level CD and layout IoU as the number of views increases. 
These results show that our method is more stable in exploiting increasing view coverage via reliability-aware pose reasoning.

\noindent
\textbf{Qualitative comparison.}
Fig. \ref{fig:qualitative_midi} and \ref{fig:qualitative_arsg} show qualitative comparisons on MIDI and ARSG, respectively. 
Our method produces more accurate object orientations and spatial arrangements, resulting in fewer misplaced objects and more coherent scene layouts.

\subsection{Ablation Studies}
\label{sec:ablation}

\begin{figure}[t]
\centering
\begin{minipage}[t]{0.48\columnwidth}
\vspace{0pt}
\centering

\captionof{table}{
\textbf{Ablation of pose routing.}
Pose guidance and multi-view reconciliation are disabled.
}
\label{tab:routing_ablation}
\small
\setlength{\tabcolsep}{3.5pt}
\renewcommand{\arraystretch}{1.08}

\vspace{-3mm}
\resizebox{\linewidth}{!}{
\begin{tabular}{lccc}
\toprule
Routing Strategy
& $\mathrm{CD}_{S}\downarrow$
& $\mathrm{FS}_{S}\uparrow$
& IoU$\uparrow$ \\
\midrule

Reference View
& 0.083 & 87.84 & 0.597 \\

Largest Mask
& 0.075 & 89.64 & 0.611 \\

Max Frame Completeness $F_v$
& 0.079 & 88.12 & 0.598 \\

Max Visible Coverage $C_v$
& 0.075 & 89.94 & 0.614 \\

Max Reliability $S_v$ (Ours)
& \textbf{0.072} & \textbf{90.08} & \textbf{0.617} \\

\bottomrule
\end{tabular}
}


\captionof{table}{
\textbf{Ablation of reconciliation.}
}
\label{tab:ablation_reconcile}
\small
\setlength{\tabcolsep}{4.0pt}
\renewcommand{\arraystretch}{1.08}

\resizebox{\linewidth}{!}{
\begin{tabular}{
>{\raggedright\arraybackslash}m{2.4cm}
*{3}{>{\centering\arraybackslash}m{0.9cm}}
}
\toprule
Variant
& $\mathrm{CD}_{S}\downarrow$
& $\mathrm{FS}_{S}\uparrow$
& IoU$\uparrow$ \\
\midrule

Object Only
& 0.062 & 86.21 & 0.651 \\

Object \& Camera
& \textbf{0.055} & \textbf{87.08} & \textbf{0.662} \\

\bottomrule
\end{tabular}
}

\end{minipage}
\hfill
\begin{minipage}[t]{0.50\columnwidth}
\vspace{0pt}
\centering

\includegraphics[width=\linewidth]{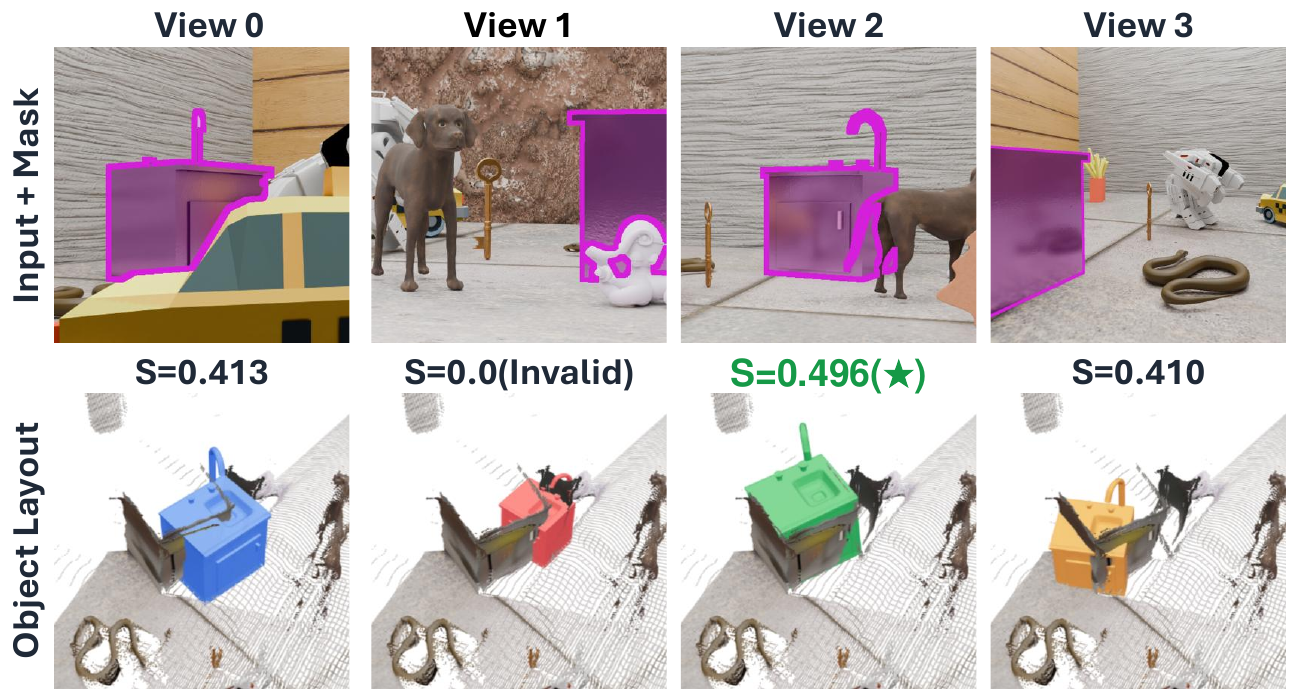}

\vspace{-2mm}
\captionof{figure}{
\textbf{Analysis of reliability-aware pose routing.}
Boundary truncation in Views~1 and~3 yields unreliable pose hypotheses. Our reliability score favors View~2, whose placement is more consistent with the scene layout.
}
\label{fig:pose_routing}
\end{minipage}
\vspace{-3mm}
\end{figure}

Table \ref{tab:ablation_main} shows that the full model achieves superior results, while removing any stage consistently degrades scene-level reconstruction and layout accuracy, confirming the complementary roles of the three components. 
Removing Guide indicates that geometry-grounded generation provides better pose initialization, while the drop without Route highlights the benefit of selecting geometrically reliable pose hypotheses. 
Moreover, the IoU gain shows the effectiveness of reconciliation in enforcing coherent scene-level placement. 
\begin{wraptable}{r}{0.55\columnwidth}
\vspace{-3mm}
\centering
\caption{\textbf{Component ablation of our method.}}
\label{tab:ablation_main}
\vspace{-2mm}
\small
\setlength{\tabcolsep}{4.5pt}
\renewcommand{\arraystretch}{1.05}

\resizebox{\linewidth}{!}{
\begin{tabular}{
l
ccccc
}
\toprule
Variant
& $\mathrm{CD}_{S}\downarrow$
& $\mathrm{FS}_{S}\uparrow$
& $\mathrm{CD}_{O}\downarrow$
& $\mathrm{FS}_{O}\uparrow$
& IoU$\uparrow$ \\
\midrule

Baseline
& 0.083 & 87.84 & 0.020 & 96.76 & 0.597 \\

w/o Guide
& 0.073 & 91.47 & 0.018 & 97.58 & 0.677 \\

w/o Route
& 0.071 & 90.63 & 0.019 & 97.10 & 0.669 \\

w/o Reconcile
& 0.072 & 90.20 & \textbf{0.014} & 97.50 & 0.617 \\

\midrule
\textbf{Ours}
& \textbf{0.069}
& \textbf{91.73}
& 0.016
& \textbf{97.66}
& \textbf{0.684} \\
\bottomrule
\end{tabular}
}
\vspace{-3mm}
\end{wraptable}
Table \ref{tab:routing_ablation} shows that pose reliability cannot be adequately captured by mask size or a single geometric cue alone.
Frame completeness and visible coverage capture complementary aspects of geometric observability, and their combination in $S_v$ enables more reliable pose selection across views. 
Fig. \ref{fig:pose_routing} further illustrates that boundary-truncated observations yield less reliable hypotheses, while $S_v$ favors a placement more consistent with the scene layout. 
Table \ref{tab:ablation_reconcile} shows that jointly refining object placements and camera poses improves scene-level reconstruction and layout accuracy. 

\section{Conclusion}
\label{sec:conclusion}

We presented \textsc{SPOON}, a scene-level pose reasoning framework for compositional 3D scene generation. 
Rather than treating object poses as independent predictions from local observations, \textsc{SPOON} uses reconstructed multi-view geometric cues as a shared scene reference to guide pose generation, route reliable pose hypotheses, and reconcile object--camera configurations through the proposed \emph{Guide--Route--Reconcile} paradigm.
This formulation turns independently inferred object--camera pose hypotheses into a coherent scene configuration while preserving the strong object priors of 3D generators. 
Experiments on ARSG-110K and MIDI-3D-Front demonstrate consistent improvements in object placement and scene composition quality across different numbers of input views.  

\clearpage
\bibliography{main}
\bibliographystyle{iclr2027_conference}


\clearpage
\appendix
\section{Appendix}

In this appendix, we first provide further methodological (\ref{sec:supp_method}) and implementation details (\ref{sec:supp_implementation}). 
We then present additional analyses and qualitative results (\ref{sec:supp_results}), followed by further discussion of limitations and future directions (\ref{sec:supp_limitation}).

\subsection{The Use of Large Language Models}
We use Large Language Models (LLMs) only for minor language editing to improve grammar and readability. All methodological design, experiments, equations, and results are developed by the authors.

\subsection{Additional Method Details}
\label{sec:supp_method}

This section provides additional details of the \textsc{SPOON} inference procedure.
We first summarize the complete \emph{Guide--Route--Reconcile} pipeline in Algorithm \ref{alg:spoon_inference}, and then detail the surface-based coordinate bridge used to transfer reconstruction-derived pose into the object canonical frame.

\begin{algorithm}[h]
\caption{\textsc{SPOON} Inference}
\label{alg:spoon_inference}
\small
\begin{algorithmic}[1]
\REQUIRE Multi-view images $\{I_v\}_{v=1}^{V}$ and object masks $\{M_{o,v}\}$
\ENSURE Compositional scene $\mathcal{S}=\{(\mathcal{G}_o,T_o^{\mathrm{ow}})\}_{o=1}^{O}$

\STATE Recover point maps $\{X_v\}$, camera intrinsics $\{K_v\}$, and world-to-camera extrinsics $\{E_v\}$ using the feed-forward reconstruction model.

\FOR{each object $o$}
    \STATE Aggregate the masked reconstruction points into the world-space object surface $\mathcal{P}_{W}^{o}$.
    \STATE Initialize the shared shape latent $\mathbf{z}^{S}$ and the view-specific pose latents $\{\mathbf{z}^{P}_{v}\}_{v=1}^{V}$.
    \STATE Initialize the surface bridge as invalid.

    \FOR{each flow step $t$}
        \STATE Predict per-view shape and pose velocities $\{\mathbf{v}^{S}_{\theta,v},\mathbf{v}^{P}_{\theta,v}\}_{v=1}^{V}$.
        \STATE Fuse $\{\mathbf{v}^{S}_{\theta,v}\}_{v=1}^{V}$ and update the shared shape trajectory.

        \IF{geometry guidance is activated}
            \IF{the surface bridge has not been initialized}
                \STATE Recover the predicted clean shape endpoint and sample the canonical surface $\mathcal{P}_{C}^{o}$.
                \STATE Estimate the surface-based coordinate bridge $\mathcal{B}^{o}_{W\rightarrow C}$ between $\mathcal{P}_{W}^{o}$ and $\mathcal{P}_{C}^{o}$.
                \STATE Validate and cache $\mathcal{B}^{o}_{W\rightarrow C}$ if the geometric alignment is reliable.
            \ENDIF
            \IF{$\mathcal{B}^{o}_{W\rightarrow C}$ is valid}
                \STATE Transfer the reconstruction-derived camera poses into the object canonical frame and construct geometry-derived pose endpoints $\{\mathbf{z}^{P,*}_{1,v}\}_{v=1}^{V}$.
                \STATE Apply flow-consistent endpoint guidance to the view-specific pose trajectories.
            \ELSE
                \STATE Apply the original pose-flow update.
            \ENDIF
        \ELSE
            \STATE Apply the original pose-flow update.
        \ENDIF
    \ENDFOR

    \STATE Decode the canonical object representation $\mathcal{G}_o$ and the view-specific pose hypotheses $\{T^{c}_{o,v}\}_{v=1}^{V}$.
    \STATE Reproject $\mathcal{P}_{W}^{o}$ into each view and compute the geometric reliability scores $\{S_{o,v}\}_{v=1}^{V}$.
    \STATE Select $v_o^{\star}=\arg\max_v S_{o,v}$ and initialize
    $T_o^{\mathrm{ow}}=E_{v_o^{\star}}^{-1}T^{c}_{o,v_o^{\star}}$.
\ENDFOR

\STATE Jointly refine the object transformations $\{T_o^{\mathrm{ow}}\}$ and camera poses $\{E_v\}$ using differentiable multi-view rendering.
\STATE \textbf{return} $\mathcal{S}=\{(\mathcal{G}_o,T_o^{\mathrm{ow}})\}_{o=1}^{O}$.
\end{algorithmic}
\end{algorithm}

\textbf{Overall Inference Procedure.}  
Given multi-view images and their object instance masks, \textsc{SPOON} first recovers a shared geometric reconstruction, including point maps and camera parameters.
For each object, during flow matching sampling, multi-view shape velocities are fused into a shared canonical shape trajectory, while the view-specific pose trajectories are geometrically guided once a sufficiently stable canonical surface becomes available.
After generation, the view-specific pose hypotheses are routed according to their geometric reliability to initialize the object placements in the shared world frame. 
Finally, object transformations and camera poses are jointly reconciled through differentiable multi-view rendering.

\textbf{Surface-Based Coordinate Bridge.} 
We provide additional details on the surface alignment used to estimate the coordinate bridge in Sec. \ref{sec:pose_guidance}. 
Given the world-space and canonical object surfaces $\mathcal{P}_{W}$ and $\mathcal{P}_{C}$ defined in the main paper, we independently center and isotropically normalize the two point sets to remove translation and global scale. 
After aligning the up axes of the reconstruction and canonical frames, we denote the resulting normalized surfaces by $\bar{\mathcal{P}}_{W}$ and $\bar{\mathcal{P}}_{C}$.
The remaining rotational ambiguity is reduced to a yaw rotation, which we estimate by:
\begin{equation}
\theta^{*} = \operatorname*{arg\,min}_{\theta}
\operatorname{TrimRMS}_{\bar{\mathbf{q}}\in\bar{\mathcal{P}}_{C}}
\left[
\min_{\bar{\mathbf{p}}\in\bar{\mathcal{P}}_{W}}
\left\|
\mathbf{R}_{\mathrm{yaw}}(\theta)^{\top}\bar{\mathbf{q}} - \bar{\mathbf{p}}
\right\|_{2}
\right].
\label{eq:supp_yaw_objective}
\end{equation}
We solve this objective using a global coarse-to-fine yaw search over the aggregated multi-view geometry. The resulting rotation is shared across all views of the same object, thereby preserving their relative camera configuration.  
The estimated yaw defines the surface bridge $\mathbf{B}^{o}_{W\rightarrow C}$ from the reconstruction world frame to the object canonical frame.
Given the reconstruction-derived world-to-camera transformation $\mathbf{E}_v$, the corresponding geometry-derived object-to-camera transformation is obtained as:
\begin{equation}
\mathbf{T}^{c,*}_{o,v}
=
\mathbf{E}_{v}
\left(\mathbf{B}^{o}_{W\rightarrow C}\right)^{-1}.
\label{eq:supp_bridge_pose}
\end{equation}
where $\mathbf{T}^{c,*}_{o,v}$ is converted to the pose parameterization of \textsc{SAM3D}, yielding $\boldsymbol{\pi}^{*}_{v}$ used in Eq. \ref{eq:geometry_pose_endpoint}. 
To improve robustness against partial observations and reconstruction outliers, we compute \(\operatorname{TrimRMS}\) using the lowest-error fraction \(\tau_{\mathrm{trim}}\) of nearest-neighbor residuals, with \(\tau_{\mathrm{trim}}=0.85\) across all experiments.
Although the reconstructed surface may only partially cover the complete canonical object, aggregating observations across multiple views provides broader geometric support than any individual view.
The normalized alignment is used only to estimate the remaining rotational correspondence between the two frames, while TrimRMS reduces the influence of unmatched regions and reconstruction outliers.
The estimated bridge is accepted only when the normalized fitting RMSE is below 0.15 and at least four valid views provide sufficient geometric support.
Otherwise, geometry guidance is disabled, and the original pose flow is used. Once accepted, the bridge is cached and reused for the remaining guided sampling steps.

\subsection{Implementation Details}
\label{sec:supp_implementation}
For geometric pose guidance, we set the guidance strength to $\alpha_P=0.8$ and activate guidance from $t=0.6$ onward, applying it at every subsequent sampling step. 
These two settings are selected based on preliminary experiments. 
For reliability-aware pose routing, we set $\alpha=2$ in Eq. \eqref{eq:reliability_score} and apply a completeness threshold $\tau_f=0.5$ to exclude severely truncated observations. 
For the reconciliation stage, we jointly optimize the object transformations and camera poses using AdamW, with learning rates of $3\times10^{-3}$ and $1\times10^{-4}$, respectively. 
The loss is a combination of photometric loss and depth loss. Since we only use depth loss as a regularization term, the depth loss is formulated as $\left\| 
M \odot \left( \exp(-d)-\exp(-\hat d) \right) 
\right\|_2^2
$, where $d$ is the depth map estimated by 3D reconstruction models, $\hat d$ is the rendered depth map, $M$ is the object mask. 

We evaluate computational efficiency on a single NVIDIA H100 GPU using scenes containing five objects under the same input setting. 
With four-view input, MV-\textsc{SAM3D} requires 195s per five-object scene, while the complete \textsc{SPOON} pipeline takes 326.5s. 
The proposed geometry guidance and pose routing introduce only modest additional costs of 10s and 1.5s, respectively, while the iterative reconciliation accounts for the remaining 120s.

We use the object instance masks provided by the benchmark datasets for all methods requiring object-level observations.
For MV-SAM3D, we follow its official inference configuration, with both layout injection during generation and post-generation object-pose refinement enabled.
Both MV-SAM3D and our method place generated objects in the world coordinate system recovered by VGGT-$\Omega$, whose global gauge generally differs from that of the ground-truth scene.
For evaluation, we estimate a single global $\mathrm{Sim}(3)$ transformation by aligning the reconstructed camera centers with the ground-truth camera centers and apply it uniformly to the entire predicted scene.
For MV-SAM3D, this transformation is estimated from the initial camera poses predicted by VGGT-$\Omega$, whereas for our method it is estimated from the camera poses after multi-view reconciliation.
No additional per-object alignment is applied for either scene-level or object-level evaluation.
For single-view baselines, where multi-view camera-center alignment is unavailable, we follow the coordinate and scale calibration protocols of their official implementations.
For example, 3D-Fixer performs in-place completion in the scene frame defined by its fragmented geometric input, with the scene scale calibrated according to its released evaluation pipeline.

\subsection{Additional Results} 
\label{sec:supp_results}

We provide additional analyses and qualitative results to further validate the proposed method. 
We first examine the effect of geometry-guided pose generation and the sensitivity of the reliability-aware routing strategy to the exponent $\alpha$.  
We then present additional qualitative comparisons on the MIDI-3D-Front and ARSG-110K test sets to demonstrate the consistency of the improvements across diverse scenes.

\begin{figure}[h]
    \centering
    \includegraphics[width=.95\textwidth]{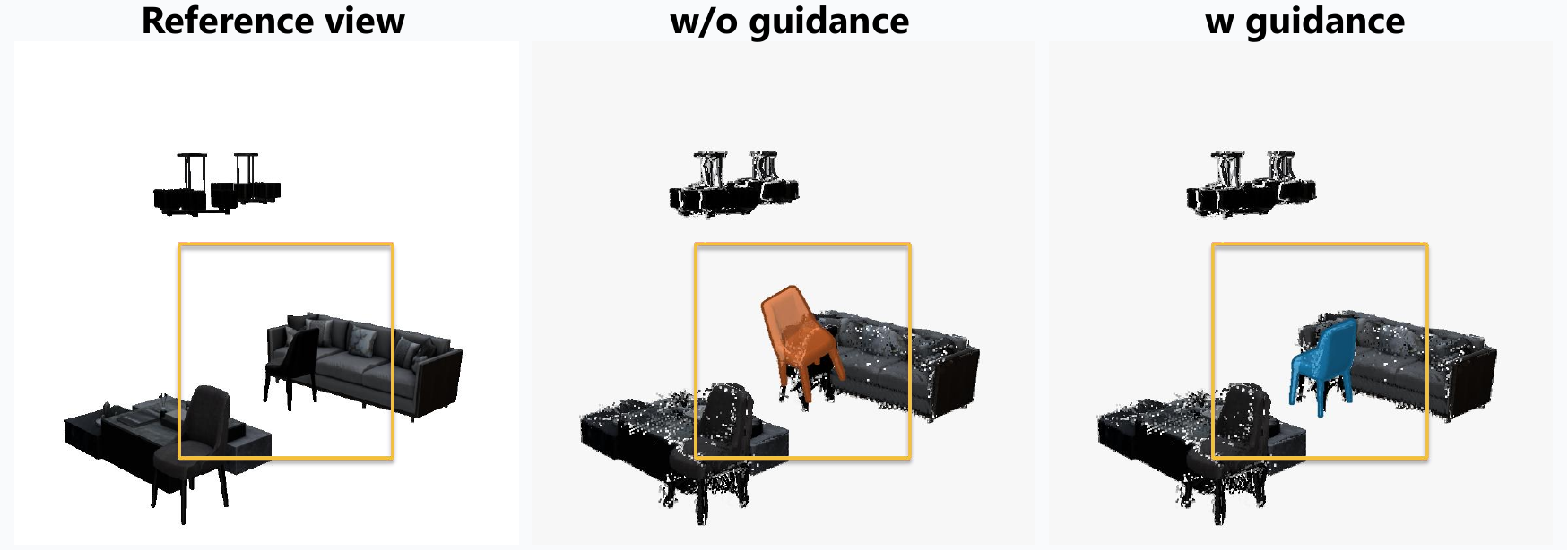}
    \vspace{-6mm}
    \caption{
    \textbf{Visualization of pose generation.}
    The highlighted object exhibits an inaccurate orientation without guidance, while pose guidance steers the generated pose toward an orientation more consistent with the object layout. 
    }
    \label{fig:vis_guidance}
    \vspace{-2mm}
\end{figure}

\textbf{Effect of pose guidance.}
Fig. \ref{fig:vis_guidance} qualitatively demonstrates the effect of geometry-guided pose generation.
Without guidance, the generated objects can exhibit noticeable orientation errors, whereas geometry guidance steers the pose toward a configuration more consistent with the observed scene layout.
This qualitatively demonstrates that the reconstructed geometric cues provide effective guidance for improving pose generation.

\begin{table}[h]
\centering
\caption{
{Sensitivity analysis of the routing exponent $\alpha$ on MIDI with four input views, with multi-view reconciliation disabled.}
}
\label{tab:alpha_sensitivity}
\small
\setlength{\tabcolsep}{6.0pt}
\renewcommand{\arraystretch}{1.08}
\vspace{-3mm}
\resizebox{.3\linewidth}{!}{
\begin{tabular}{cccc}
\toprule
$\alpha$
& $\mathrm{CD}_{S}\downarrow$
& $\mathrm{FS}_{S}\uparrow$
& IoU$\uparrow$ \\
\midrule
1   & 0.072 & 90.08 & 0.619 \\
2   & 0.072 & 90.20 & 0.617 \\
3   & 0.072 & 90.07 & 0.618 \\
\bottomrule
\end{tabular}
}
\vspace{-2mm}
\end{table}

\textbf{Sensitivity to the routing exponent.}
Table \ref{tab:alpha_sensitivity} shows that the routing strategy is insensitive to the choice of $\alpha$. 
Across $\alpha\in\{1,2,3\}$, all three metrics remain nearly unchanged, indicating that the proposed reliability score is robust to moderate variations in the completeness weighting.

\begin{figure*}[t]
    \centering
    \includegraphics[width=\textwidth]{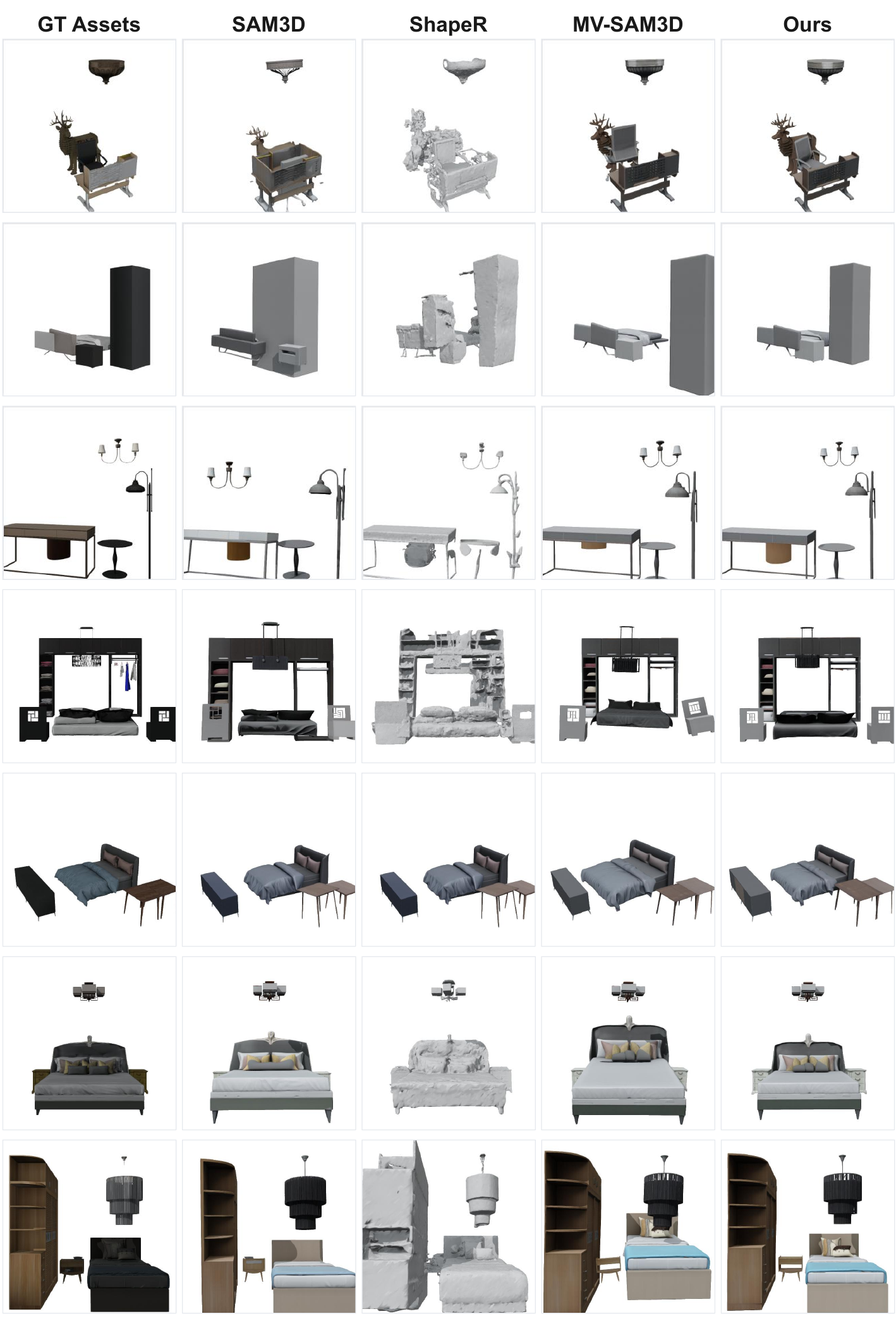}
    \vspace{-6mm}
    \caption{
    \textbf{Qualitative comparison on the MIDI-3D-Front test set.}
    }
    \label{fig:qualitative_midi_supp}
    \vspace{-2mm}
\end{figure*}

\begin{figure*}[t]
    \centering
    \includegraphics[width=\textwidth]{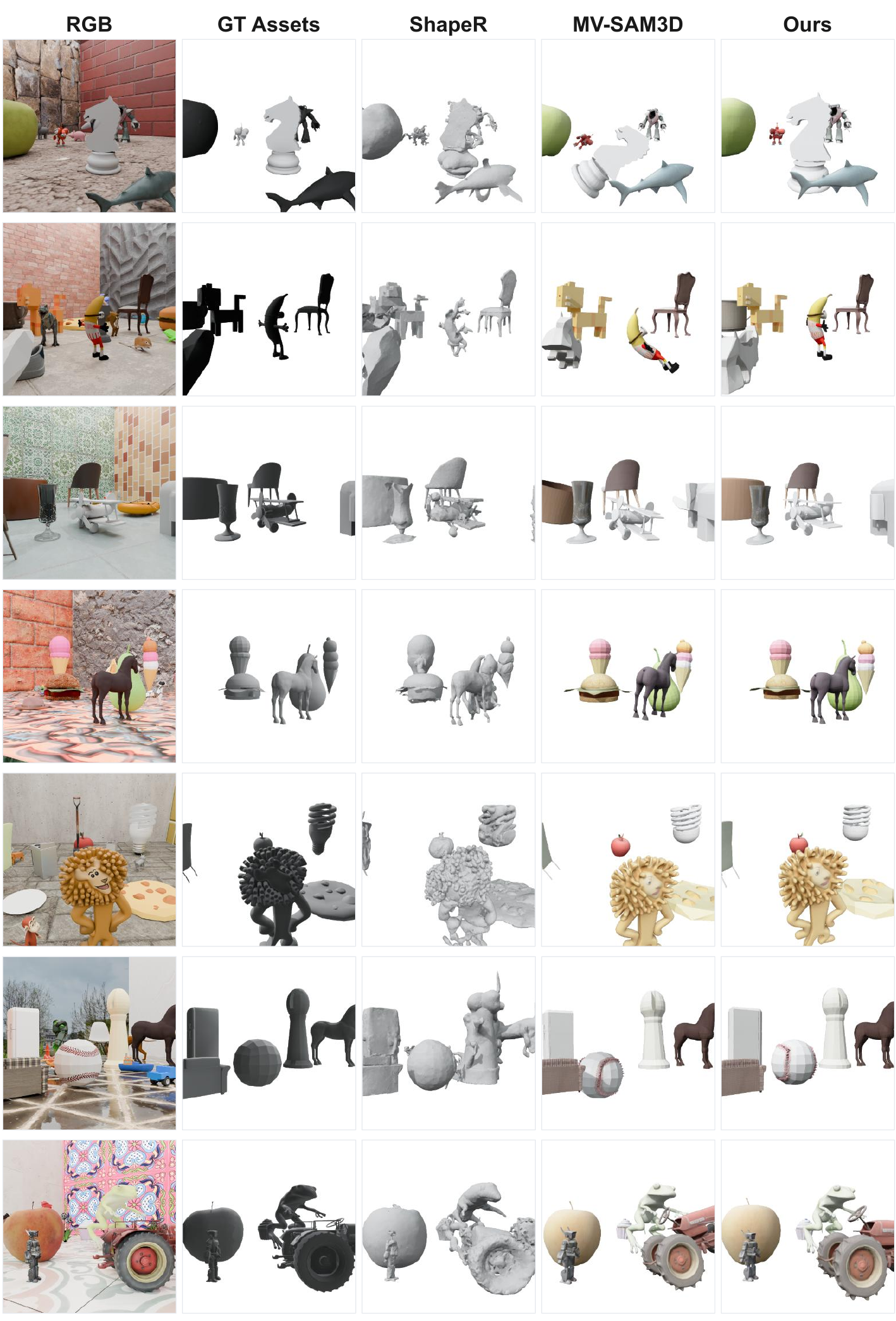}
    \vspace{-6mm}
    \caption{
    \textbf{Qualitative comparison on the ARSG-110K test set.}
    }
    \label{fig:qualitative_arsg_supp}
    \vspace{-2mm}
\end{figure*}

\textbf{Additional qualitative comparisons.}
Figs. \ref{fig:qualitative_midi_supp} and \ref{fig:qualitative_arsg_supp} provide additional qualitative comparisons on the MIDI-3D-Front and ARSG-110K test sets, respectively.
Across diverse scenes, our method produces object configurations with more consistent orientations and relative placements, resulting in scene layouts that better agree with the reference observations.
These results further demonstrate the benefit of geometry-grounded pose reasoning for improving scene-level spatial coherence under multi-view observations.

\subsection{Limitation and Future Work}
\label{sec:supp_limitation}
Although \textsc{SPOON} improves scene-level coherence, individual objects are still generated independently rather than through an explicit multi-instance generative process. 
Extending the current object-wise formulation toward joint multi-instance generation could enable stronger coupling between object generation and scene-level spatial organization. 
In addition, the effectiveness of geometry-grounded reasoning depends on the quality of the underlying multi-view reconstruction and may be affected by inaccurate or incomplete geometry. 


\end{document}